%% file: main.tex
\providecommand{\main}{.}
\documentclass[runningheads]{llncs}
\usepackage[T1]{fontenc}
\usepackage{graphicx}
\usepackage{color}
\usepackage[english]{babel}
\usepackage[square,numbers]{natbib}
\usepackage{subfiles}
\usepackage{hyperref}
\usepackage{todonotes}
\usepackage{xcolor}
\usepackage{amsfonts}
\usepackage{amssymb}
\usepackage{amsmath}
\usepackage{graphicx,subfigure}
\usepackage{standalone}
\usepackage{stackengine,scalerel}

\usepackage[T1]{fontenc}
\usepackage{pgfplots}
\pgfplotsset{compat=1.16}
\usetikzlibrary{intersections}
\usepackage{mwe}
\usepackage{tikz,fp}
\usepackage{makecell}
\usepackage{algorithm}
\usepackage{algpseudocode}
\usepackage{amsbsy}
\usepackage{booktabs,tabularx}

\usepackage{float}
\usepackage{authblk}
\usepackage{bookmark}
\usepackage{wrapfig}
\usepackage{graphicx}
\usepackage{tablefootnote}
\usepackage{multirow}

\graphicspath{{tikz/}}

\newcommand\seq[1]{\boldsymbol{#1}}

\usetikzlibrary{matrix,calc,shapes}
\usetikzlibrary{bayesnet}
\usetikzlibrary{calc}

\usetikzlibrary{decorations.text}

\tikzset{
    partial ellipse/.style args={#1:#2:#3}{
        insert path={+ (#1:#3) arc (#1:#2:#3)}
    }
}

\tikzset{fit margins/.style={/tikz/afit/.cd,#1,
    /tikz/.cd,
    inner xsep=\pgfkeysvalueof{/tikz/afit/left}+\pgfkeysvalueof{/tikz/afit/right},
    inner ysep=\pgfkeysvalueof{/tikz/afit/top}+\pgfkeysvalueof{/tikz/afit/bottom},
    xshift=-\pgfkeysvalueof{/tikz/afit/left}+\pgfkeysvalueof{/tikz/afit/right},
    yshift=-\pgfkeysvalueof{/tikz/afit/bottom}+\pgfkeysvalueof{/tikz/afit/top}},
    afit/.cd,left/.initial=2pt,right/.initial=2pt,bottom/.initial=2pt,top/.initial=2pt}

\usepackage{xcolor,colortbl}
\definecolor{Gray}{gray}{0.85}
\definecolor{LightCyan}{rgb}{0.88,1,1}
\definecolor{LightBlue}{rgb}{0.78,0.86,0.93}
\definecolor{LightGreen}{rgb}{0.75,0.89,0.75}
\definecolor{Blue}{rgb}{0.12,0.47,0.71}
\definecolor{LightOrange}{rgb}{1,0.87,0.76}
\definecolor{Orange}{rgb}{0.97,0.55,0.17}
\newcolumntype{a}{>{\columncolor{Gray}}c}
\newcolumntype{b}{>{\columncolor{LightCyan}}c}
\newcommand{\myblue}[1]{\textcolor{Blue}{#1}}
\newcommand{\myorange}[1]{\textcolor{Orange}{#1}}

\usepackage{todonotes}
\usepackage{lipsum} 

\newboolean{anonymous}
\setboolean{anonymous}{false}

\algtext*{EndWhile}
\algtext*{EndIf}
\begin{document}

%
%
%
\title{Online learning in motion modeling for intra-interventional image sequences}
\ifthenelse
{\boolean{anonymous}}
{
    \author{Author~1 \inst{1,2}\orcidID{****-****-****-****} \and
    Author~2\inst{1}\orcidID{****-****-****-****} \and
    Author~3\inst{2}\orcidID{****-****-****-****} \and
    Author~4\inst{1}\orcidID{****-****-****-****}}    

    \authorrunning{Author 1 et al.}
%
%
\institute{****, ****, **** \\
\email{**@******.***} \and
****, ****, ****\\
\email{**@******.***}}
}
{
\author{Niklas Gunnarsson\inst{1,2}\orcidID{0000-0002-9013-949X} \and
Jens~Sj\"olund\inst{1}\orcidID{0000-0002-9099-3522} \and
Peter~Kimstrand\inst{2}\orcidID{0000-0001-9667-5595} \and
Thomas~B.~Sch\"on\inst{1}\orcidID{0000-0001-5183-234X}}
%
\authorrunning{N. Gunnarsson et al.}
%
\institute{Department of Information Technology, Uppsala University, Sweden \\
\email{\{firstname\}.\{surname\}@it.uu.se} \and
Elekta Instrument AB, Stockholm, Sweden\\
\email{\{firstname\}.\{surname\}@elekta.com}}
}

\maketitle              
\begin{abstract}
    Image monitoring and guidance during medical examinations can aid both diagnosis and treatment. However, the sampling frequency is often too low, which creates a need to estimate the missing images. We present a probabilistic motion model for sequential medical images, with the ability to both estimate motion between acquired images and forecast the motion ahead of time. The core is a low-dimensional temporal process based on a linear Gaussian state-space model with analytically tractable solutions for forecasting, simulation, and imputation of missing samples. The results, from two experiments on publicly available cardiac datasets, show reliable motion estimates and an improved forecasting performance using patient-specific adaptation by online learning.
\end{abstract}
\subfile{./01_Introduction.tex}
\subfile{./02_Background.tex}
\subfile{./03_Method.tex}
\subfile{./04_Experiment.tex}
\subfile{./05_Discussion.tex}

\begin{credits}

\subsubsection{\ackname} \ifthenelse{\boolean{anonymous}}{***}{This research was partially supported by the \emph{Wallenberg AI, Autonomous Systems and Software Program (WASP)} funded by Knut and Alice Wallenberg Foundation.}

\subsubsection{\discintname}
The authors have no competing interests to declare that are relevant to the content of this article.
\end{credits}
%
%
%
\newpage
\bibliography{ref}

\newpage
\subfile{./SupplementaryFile.tex}


\end{document}

%% file: 01_Introduction.tex
\section{Introduction}

Sequential imaging during medical interventions, so-called intra-interventional imaging, appears in several medical examinations. In cardiology, diagnostic decisions may be supported by cardiac ultrasound or cardiac MRI by acquiring images of the heart over one or several cardiac cycles~\cite{angelini2001lv}. In MR-guided radiotherapy~\cite{raaymakers2009integrating}, 2D cine MRI is used to monitor moving tumors and organs at risk during ongoing treatment sessions. This enables controlling and adapting the treatment beam~\cite{keall2022integrated}.

A common desire is to identify the anatomical motion from a static reference image to each subsequent image in the temporal sequence. This enables transferring segmentations (e.g. organs) identified in the reference image and estimating their location in the sequence. Finding the corresponding deformation field is the main goal of motion estimation. Of particular interest are diffeomorphic deformations, which are topology-preserving and ensure one-to-one mapping between the pixels/voxels in the two images. Examples of conventional diffeomorphic image registration methods are Large Deformation Diffeomorphic Metric Mapping~\cite{beg2005computing} and symmetric normalization~\cite{avants2008symmetric}. Recently, deep learning image registration methods~\cite{krebs2019learning,ye2023sequencemorph} have shown fast and accurate performance in motion detection and organ tracking by removing the iterative optimization procedure from inference time and solving tasks in nearly real-time. However, image registration methods do not consider the sequential nature of an image sequence and estimate the motion using one image pair at a time. 

With sequential images, an interesting research question is to model the temporal sequence from the data. We refer to this as motion modeling -- a model with the possibility of estimating the motion at the previous, current, and future times. We present a diffeomorphic motion model suitable for intra-interventional medical image sequences. For this, we define and model a temporal process in a low-dimensional latent space with the possibility to impute and forecast missing samples in the sequence. Furthermore, our model is the first, to the best of our knowledge, to support online learning of the temporal model which makes it suitable for real-time scenarios.

\section{Related work} 
Motion modeling and real-time analysis of intra-interventional medical images is a relatively new research direction. The literature shows that the most common motion to analyze is cyclic patterns like cardiac or respiratory motion~\cite{chen2020deep,paganelli2018mri}. A general approach is to embed the image data into a lower dimensional space and model the temporal process in this domain. Romaguera et al. \cite{romaguera2020prediction} present a forecasting approach where they suggest a convolutional LSTM to extrapolate the temporal process in the latent dimension. Extension of their work includes a forecasting 4D motion (3D + time) given 2D intra-interventional images using a probabilistic setting~\cite{romaguera2021probabilistic}. Krebs et al.~\cite{krebs2021learning} proposed a more general probabilistic motion model. Their model relies on a conditional variational autoencoder, where they approximate the posterior distribution in the latent dimension using a temporal convolutional neural network. During training, they minimize the Kullback–Leibler divergence between their approximate posterior distribution and a known Gaussian process prior. Missing samples in the sequence are then replaced with samples from this prior. Their work shows reliable diffeomorphic estimates of the displacement field with imputing and forecasting possibility. However, they are limited to image sequences of fixed length. To overcome this, \ifthenelse{\boolean{anonymous}}{***}{Gunnarsson et al.}~\cite{gunnarsson2022unsupervised} modeled the low dimensional temporal process using a linear Gaussian state space model, i.e., a first-order Markov process. The work we present here is a further development of this model, including support for online learning and architectural improvements.

%% file: 02_Background.tex
\section{Background - Linear Gaussian state space model}
A linear Gaussian state space model~(LG-SSM) is a linear representation of a state space model. The model defines a first-order Markov process for a dynamic state variable $z_t \in \mathbb{R}^p$ followed by a transmission operation between the state variable and an observed variable $x_t \in \mathbb{R}^q$, i.e. 
\begin{equation}
    \begin{aligned}
 z_t | z_{t-1} \sim \mathcal{N}(z_t | Az_{t-1}, Q), \quad  x_t | z_t \sim \mathcal{N}(x_t | Cz_t, R), \quad z_0 \sim \mathcal{N}(z_0 | \mu_0, P_0), \label{eq:lgssm}
\end{aligned}
\end{equation}
where $A \in \mathbb{R}^{p \times p}$ and $C \in \mathbb{R}^{q \times p}$ denote the state and observation matrix, respectively, $Q \in \mathbb{R}^{p \times p}$ and $R \in \mathbb{R}^{q \times q}$ denote covariance matrices for uncertainties and $\mu_0$, $P_0$ is the initial values of the state process. Besides that, LG-SSMs are beneficial since the state prediction, $z_{t+k} \mid z_{t}$, and smoothing, $z_{t-k} \mid z_{t}$, $k>0$, are analytically tractable using i.e. Kalman filtering~\cite{kalman1960new} and RTS smoothing~\cite{rauch1965maximum}. Lately, parameter-estimated LG-SSM has shown impressive results in long-range sequence modeling tasks, outperforming recent methods like RNNs, CNNs, and Transformers\cite{gu2021efficiently}.

To reduce the computational complexity of high-dimensional sequences $\seq{y} = [y_1, \dots, y_T]$, like videos, Fraccaro et al. \cite{fraccaro2017disentangled} proposed a probabilistic dynamical model that embeds the sequence into a lower-dimensional space where it is represented as an LG-SSM.

%% file: 03_Method.tex
\section{Method}
\begin{figure}
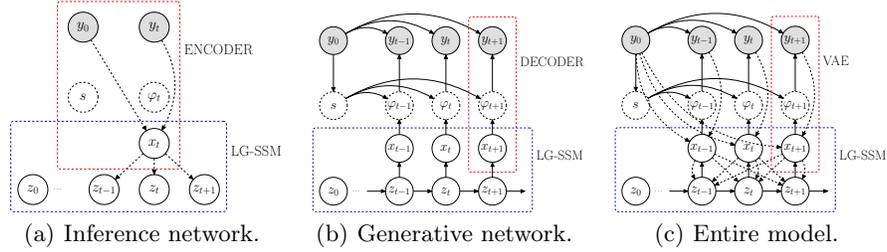

     \hfill
     \subfigure[Inference network.]{
     \resizebox{0.30\linewidth}{!}{\subfile{\main/tikz/model_inference}}\label{fig:graph_i}}
     \hfill
     \subfigure[Generative network.]{
     \resizebox{0.30\linewidth}{!}{\subfile{\main/tikz/model_generative}}\label{fig:graph_g}}
     \hfill
     \subfigure[Entire model.]{
     \resizebox{0.30\linewidth}{!}{\subfile{\main/tikz/model_g}}\label{fig:graph}}
     \hfill
     \caption{An observation $y_t$ is downsampled to the temporal process \subref{fig:graph_i}. The spatial transformation is generated given $y_0$ and the low-dimensional motion model~\subref{fig:graph_g}. \subref{fig:graph} visualizes the entire model.}
     \label{fig:model_gi}
 \end{figure}

Given the data $\{(y_0, \seq{y})^{(i)}\}_{i=1}^n$ of static reference images $y_0$ and time sequences $\seq{y} = [y_1, \dots y_T]$ our goal is to first model the spatiotemporal changes and then use this model to reconstruct and generate samples at other times. For this, we explain the spatiotemporal changes as the spatial transformation $\varphi_t$ from the static reference image to each time step $t$ in the sequence such that $y_t \approx y_0 \circ \varphi_t$~\cite{gunnarsson2022unsupervised}. To include spatial information in the transformation, like contour information and description of shapes, but still limit the temporal process to the most significant temporal changes, we estimate $\varphi_t$ based on the temporal process and spatial information $s$ given the static reference image $y_0$, i.e,
\begin{align}
    \varphi_t = g_{\theta_{\text{g}}}(x_t, s), \quad s = f_{\theta_{\text{s}}}(y_0),
\end{align}
where $x_t$ is a low-dimensional variable at time $t$. By doing this, we can separate the temporal changes and characteristic features from the images within the sequence. Since the image process may be incomplete due to e.g. missing samples, we define the spatial information based on the static reference image $y_0$ only. The spatial transformation $\varphi_t$ is a function of $x_t$ and $s$ (and $s$ is a function of $y_0$), we can parameterize $p_\theta(\seq{y} \mid y_0, \seq{x}) = \prod_{t=1}^T p_\theta(y_t \mid y_0, x_t)$ with a generative network with parameters $\theta = \{\theta_{\text{g}}, \theta_{\text{s}}\}$ and model the likelihood $p_\theta(y_t \mid y_0, x_t) = p_\theta(y_t \mid y_0, \varphi_t)$ as any computable continuous distribution in $\theta$. Furthermore, to estimate missing samples in the sequence we model the temporal process in the lower dimension using an LG-SSM, driven by the state variables $z_t$ and with parameters $\gamma = \{A, Q, C, R, \mu_0, \Sigma_0\}$. Finally, given an approximate posterior $q(\seq{x},\seq{z} \mid y_0, \seq{y}) = q_\phi(\seq{x} \mid y_0, \seq{y}) p_\gamma(\seq{z} \mid \seq{x})$ an evidence lower bound (ELBO) can be derived as
 \begin{equation}
     \begin{aligned}
         \log p(\seq{y} \mid y_0) 
         & \geq  \mathbb{E}_{q_\phi(\seq{x} \mid y_0, \seq{y})} 
         \Big[\log \dfrac{
             p_\theta(\seq{y} \mid y_0, \seq{\varphi})
         }{
             q_\phi(\seq{x} \mid y_0, \seq{y})
         }
         + \mathbb{E}_{p_\gamma(\seq{z} \mid \seq{x})} 
         \Big[\log \dfrac{
             p_\gamma(\seq{x}, \seq{z})
         }{
             p_\gamma(\seq{z} \mid \seq{x})
         }\Big] \Big], 
         \label{eq:elbo_final}
     \end{aligned}
 \end{equation}
 where $p_\gamma(\seq{x}, \seq{z})$ and $p_\gamma(\seq{z} \mid \seq{x})$ are both analytical tractable using Kalman filtering and RTS smoothing. During the training process we maximize the approximate ELBO by sampling $\tilde{\seq{x}} \sim q_\phi(\seq{x} \mid y_0, \seq{y})$ and $\tilde{\seq{z}} \sim p_\gamma(\seq{z} \mid \tilde{\seq{x}})$ and update the parameters of the inference network ($\phi$), the LG-SSM ($\gamma$) and the generative network ($\theta$) simultaneously. For a complete derivation of the ELBO, we refer to supplemental material. A schematic overview of our probabilistic model given the observed variables $\seq{y}, y_0$ and unobserved variables $\seq{x}, \seq{z}$ is shown in Fig.~\ref{fig:model_gi}.
\subsection{Online learning}
\begin{figure}
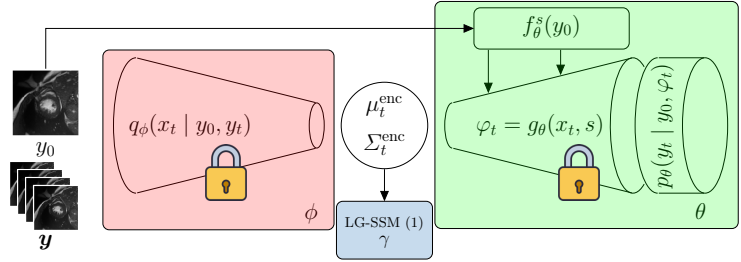

    \centering
    \resizebox{0.8\linewidth}{!}{%
    \subfile{\main/tikz/model}}
    \caption{During online-learning we fix the parameters of the encoder ($\phi$) and the decoder ($\theta$), and only update the parameters of the LG-SSM ($\gamma$).}%
    \label{fig:illustration}
\end{figure}

To adapt the model for individual patient motion, we propose a fast online learning procedure that operates on the motion model only. This means we only focus on the LG-SSM parameters $\gamma$ and keep the inference and generative network parameters $\{\phi, \theta\}$ fixed (shown in Fig. \ref{fig:illustration}). To update the LG-SSM parameters, we iteratively maximize the exact marginal log-likelihood for the $N$ most recent samples of the temporal process at each sampling time $t$, i.e,
\begin{align}
    \max_{\gamma_t} \log p_{\gamma_t}(x_{t-N:t}) = \max_{\gamma_t} \log \prod_{k=t-N}^{t} p_{\gamma_t}(x_{k} \mid x_{k-1}). \label{eq:online_learning}
\end{align}
This approach is based on the moving horizon estimation technique~\cite{mattingley2010real}, which is a well-established method for state estimation in real-time applications. We calculate the marginal log-likelihood using the Kalman filter and update the parameters using gradient-based optimization methods. The algorithm for our proposed online learning procedure is shown in Algorithm~\ref{alg:online_learning}.
\begin{algorithm}
    \caption{Online training}\label{alg:online_learning}
    \subfile{\main/algorithms/online_learning}
\end{algorithm}
\subsection{Implementation details}
In our implementation, the inference network and the spatial feature extraction share a similar network architecture. We downsample the data using convolutional layers with filters $[32, 32, 32, 16]$, extract the spatial features at each level, and estimate the mean and covariance of $x_t$ at the bottom level. For the LG-SSM, we use eight dimensions for $x_t$ ($p=8$) and $16$ for the state-variable $z_t$ ($q=16$) and estimate all model parameters. In the generative network, we use attention gates~\cite{oktay2018attention} to focus the temporal changes on the spatial features of the reference image at each resolution and use the same number of filters as in the inference network. To ensure diffeomorphic estimation of $\varphi_t$ we consider the output as the stationary velocity field $v_t$ and first smooth it using a Gaussian filter~\cite{krebs2019learning} and then compute the transformation numerically using four scaling-and-squaring layers~\cite{arsigny2006log}, a proven approach to obtain diffeomorphic registrations~\cite{dalca2019unsupervised,krebs2019learning}. Our implementation is publicly available at \ifthenelse{\boolean{anonymous}}{***}{\url{https://github.com/ngunnar/2D_motion_model}} and for a more detailed description, we refer to supplemental material.

%% file: tikz/model_inference.tex
\pgfmathsetmacro{\R}{1.5cm}
\begin{tikzpicture}[x=1.0cm,y=0.8cm]
    \node[obs, minimum size=\R] (y1) {\huge $y_{t}$};

    \node[const, left=of y1]  (t1_0) {}; %
    \node[obs, minimum size=\R, left= of t1_0] (y0) {\huge $y_0$};
    
    \node[latent, dashed, minimum size=\R, below= 2cm  of y0] (s) {\huge $s$};
    \node[latent, dashed, minimum size=\R, below= 2cm of y1] (phi2) {\huge $\varphi_{t}$};
    
    \node[latent, minimum size=\R, below=of phi2] (x2) {\huge $x_{t}$};

    \node[latent, minimum size=\R, below=of x2] (z2) {\huge $z_{t}$};
    \node[latent, minimum size=\R, left= of z2] (z1) {\huge $z_{t-1}$};
    \node[latent, minimum size=\R, right=of z2] (z3) {\huge $z_{t+1}$};
    \node[const, left=of z1]  (t_0) {}; %
    \node[const, right=of z3]  (t_N) {}; %
       
    \node[latent, minimum size=\R, left= of t_0] (z0) {\huge $z_0$};
    
    \path (y0) edge [dashed, -triangle 45] node {} (x2);
    
    \path (y1) edge [bend left, dashed, -triangle 45] node {} (x2);   

    \path (x2) edge [dashed, -triangle 45] node {} (z1);    
    \path (x2) edge [dashed, -triangle 45] node {} (z2);
    \path (x2) edge [dashed, -triangle 45] node {} (z3);
    
    \node (b) at ($(z0)!0.35!(z1)$) {$\dots$};

    \gate[red, rounded corners,inner sep=15pt] {vae} {(y0)(s)(y1)(x2)} {} ; %
    \gate[blue, rounded corners,inner sep=10pt] {ssm} {(z0)(t_0)(x2)(z1)(z2)(z3)} {} ; %
    
    \node[const, right=of vae, xshift=-0.8cm, yshift=1.8cm]  () {\huge ENCODER};
    \node[const, right=of ssm, xshift=-0.8cm, yshift=0.8cm]  () {\huge LG-SSM};
\end{tikzpicture}

%% file: tikz/model_generative.tex
\pgfmathsetmacro{\R}{1.5cm}
\begin{tikzpicture}[x=1.0cm,y=0.8cm]
    \node[obs, minimum size=\R] (y1) {\huge $y_{t-1}$};
    \node[obs, minimum size=\R, right=of y1] (y2) {\huge $y_{t}$};
    \node[obs, minimum size=\R, right=of y2] (y3) {\huge $y_{t+1}$};
    
    \node[const, left=of y1]  (t1_0) {}; %
    \node[obs, minimum size=\R, left=of t1_0] (y0) {\huge $y_0$};
    
    \node[latent, dashed, minimum size=\R, below= 2cm  of y0] (xfeat) {\huge $s$};
    \node[latent, dashed, minimum size=\R, below= 2cm of y1] (phi1) {\huge $\varphi_{t-1}$};
    \node[latent, dashed, minimum size=\R, below= 2cm of y2] (phi2) {\huge $\varphi_{t}$};
    \node[latent, dashed, minimum size=\R, below= 2cm of y3] (phi3) {\huge $\varphi_{t+1}$};    
    
    \node[latent, minimum size=\R, below= of phi1] (x1) {\huge $x_{t-1}$};
    \node[latent, minimum size=\R, right=of x1] (x2) {\huge $x_{t}$};
    \node[latent, minimum size=\R, right=of x2] (x3) {\huge $x_{t+1}$};    
    
    \node[latent, minimum size=\R, below= of x1] (z1) {\huge $z_{t-1}$};
    \node[latent, minimum size=\R, right=of z1] (z2) {\huge $z_{t}$};
    \node[latent, minimum size=\R, right=of z2] (z3) {\huge $z_{t+1}$};
    \node[const, left=of z1]  (t_0) {}; %
    \node[const, right=of z3]  (t_N) {}; %
    \node[latent, minimum size=\R, left= of t_0] (z0) {\huge $z_0$};
    
    \path (y0) edge [-triangle 45] node {} (s);

    \path (x1) edge [-triangle 45] node {} (phi1) ;
    \path (phi1) edge[-triangle 45] node {} (y1);
    \path (x2) edge [-triangle 45] node {} (phi2) ;
    \path (phi2) edge [-triangle 45] node {} (y2);
    \path (x3) edge [-triangle 45] node {} (phi3) ;
    \path (phi3) edge [-triangle 45] node {} (y3);

    \path (z1) edge [-triangle 45] node {} (x1) ;
    \path (z2) edge [-triangle 45] node {} (x2) ;
    \path (z3) edge [-triangle 45] node {} (x3) ;
    
    \path (t_0) edge [-triangle 45] node {} (z1) ;
    \path (z1) edge [-triangle 45] node {} (z2) ;
    \path (z2) edge [-triangle 45] node {} (z3) ;c
    \path (z3) edge [-triangle 45] node {} (t_N);
    
    \path (y0) edge [bend left, -triangle 45] node {} (y1);
    \path (y0) edge [bend left, -triangle 45] node {} (y2);
    \path (y0) edge [bend left, -triangle 45] node {} (y3);
        
    \path (xfeat) edge [bend left, -triangle 45] node {} (phi1);
    \path (xfeat) edge [bend left, -triangle 45] node {} (phi2);
    \path (xfeat) edge [bend left, -triangle 45] node {} (phi3);

    \node (b) at ($(z0)!0.35!(z1)$) {$\dots$};

    \gate[red, rounded corners,inner sep=15pt] {vae} {(y3)(x3)} {} ; %
    \gate[blue, rounded corners,inner sep=10pt] {ssm} {(z0)(t_0)(x1)(x2)(x3)(t_N)(z1)(z2)(z3)} {} ; %
    
    \node[const, right=of vae, xshift=-0.8cm, yshift=1.8cm]  () {\huge DECODER};
    \node[const, right=of ssm, xshift=-0.8cm, yshift=0.8cm]  () {\huge LG-SSM};
\end{tikzpicture}

%% file: tikz/model_g.tex
\pgfmathsetmacro{\R}{1.5cm}
\begin{tikzpicture}[x=1.0cm,y=0.8cm]
    \node[obs, minimum size=\R] (y1) {\huge $y_{t-1}$};
    \node[obs, minimum size=\R, right=of y1] (y2) {\huge $y_{t}$};
    \node[obs, minimum size=\R, right=of y2] (y3) {\huge $y_{t+1}$};
    
    \node[const, left=of y1]  (t1_0) {}; %
    \node[obs, minimum size=\R, left=of t1_0] (y0) {\huge $y_0$};
    
    \node[latent, dashed, minimum size=\R, below= 2cm  of y0] (xfeat) {\huge $s$};
    \node[latent, dashed, minimum size=\R, below= 2cm of y1] (phi1) {\huge $\varphi_{t-1}$};
    \node[latent, dashed, minimum size=\R, below= 2cm of y2] (phi2) {\huge $\varphi_{t}$};
    \node[latent, dashed, minimum size=\R, below= 2cm of y3] (phi3) {\huge $\varphi_{t+1}$};    
    
    \node[latent, minimum size=\R, below= of phi1] (x1) {\huge $x_{t-1}$};
    \node[latent, minimum size=\R, right=of x1] (x2) {\huge $x_{t}$};
    \node[latent, minimum size=\R, right=of x2] (x3) {\huge $x_{t+1}$};

    \node[latent, minimum size=\R, below= of x1] (z1) {\huge $z_{t-1}$};
    \node[latent, minimum size=\R, right=of z1] (z2) {\huge $z_{t}$};
    \node[latent, minimum size=\R, right=of z2] (z3) {\huge $z_{t+1}$};
    \node[const, left=of z1]  (t_0) {}; %
    \node[const, right=of z3]  (t_N) {}; %
    \node[latent, minimum size=\R, left= of t_0] (z0) {\huge $z_0$};
    
    \path (x1) edge [-triangle 45] node {} (phi1) ;
    \path (phi1) edge[-triangle 45] node {} (y1);
    \path (x2) edge [-triangle 45] node {} (phi2) ;
    \path (phi2) edge [-triangle 45] node {} (y2);
    \path (x3) edge [-triangle 45] node {} (phi3) ;
    \path (phi3) edge [-triangle 45] node {} (y3);

    \path (z1) edge [-triangle 45] node {} (x1) ;
    \path (z2) edge [-triangle 45] node {} (x2) ;
    \path (z3) edge [-triangle 45] node {} (x3) ;
    
    \path (t_0) edge [-triangle 45] node {} (z1) ;
    \path (z1) edge [-triangle 45] node {} (z2) ;
    \path (z2) edge [-triangle 45] node {} (z3) ;c
    \path (z3) edge [-triangle 45] node {} (t_N);
    
    \path (y0) edge [bend left, -triangle 45] node {} (y1);
    \path (y0) edge [bend left, -triangle 45] node {} (y2);
    \path (y0) edge [bend left, -triangle 45] node {} (y3);

    \path (y0) edge [bend right, dashed, -triangle 45] node {} (x1);
    \path (y0) edge [bend right, dashed, -triangle 45] node {} (x2);
    \path (y0) edge [bend right, dashed, -triangle 45] node {} (x3);
    
    \path (y1) edge [bend left, dashed, -triangle 45] node {} (x1);
    \path (y2) edge [bend left, dashed, -triangle 45] node {} (x2);
    \path (y3) edge [bend left, dashed, -triangle 45] node {} (x3);
    
    \path (y0) edge [-triangle 45] node {} (xfeat);
    \path (xfeat) edge [bend left, -triangle 45] node {} (phi1);
    \path (xfeat) edge [bend left, -triangle 45] node {} (phi2);
    \path (xfeat) edge [bend left, -triangle 45] node {} (phi3);

    \path (x1) edge [bend right, dashed, -triangle 45] node {} (z1);    
    \path (x1) edge [dashed, -triangle 45] node {} (z2);
    \path (x1) edge [dashed, -triangle 45] node {} (z3);

    \path (x2) edge [dashed, -triangle 45] node {} (z1);    
    \path (x2) edge [bend left, dashed, -triangle 45] node {} (z2);
    \path (x2) edge [dashed, -triangle 45] node {} (z3);

    \path (x3) edge [dashed, -triangle 45] node {} (z1);    
    \path (x3) edge [dashed, -triangle 45] node {} (z2);
    \path (x3) edge [bend left, dashed, -triangle 45] node {} (z3);
    
    \node (b) at ($(z0)!0.35!(z1)$) {$\dots$};

    \gate[red, rounded corners,inner sep=15pt] {vae} {(y3)(x3)} {} ; %
    \gate[blue, rounded corners,inner sep=10pt] {ssm} {(z0)(t_0)(x1)(x2)(x3)(t_N)(z1)(z2)(z3)} {} ; %
    
    \node[const, right=of vae, xshift=-0.8cm, yshift=1.8cm]  () {\huge VAE};
    \node[const, right=of ssm, xshift=-0.8cm, yshift=0.8cm]  () {\huge LG-SSM};
\end{tikzpicture}

%% file: algorithms/online_learning.tex
\begin{algorithmic}
    \Require $\gamma_0, \phi, \theta, y_0, N$
    \State $\seq{s} \gets f_\phi(y_0)$
    \State $x \gets \{\}$
    \While{$y_t$ arrives}        
        \State $\tilde{x}_t \sim q_\theta(x_t \mid y_0, y_t)$ 
        \State $x \gets (x \setminus \tilde{x}_{t-N}) \cup \tilde{x}_t$ \Comment{Data collection}
        \State $p_{\gamma_{t-1}}(x) \gets \textnormal{Kalman Filter}(x)$
        \State $\mathcal{L} \gets \log p_{\gamma_{t-1}}(x)$ \Comment{Equation (\ref{eq:online_learning})}
        \State $\gamma_{t} \gets \textnormal{update}(\gamma_{t-1}, \nabla_{\gamma_{t-1}} \mathcal{L})$
    \EndWhile
\end{algorithmic}

%% file: tikz/model.tex
\begin{tikzpicture}
    \node[label=below:{\Large $y_0$},draw=white, inner sep=0] (imgM) at (-2,2) {\includegraphics[height=1.5cm]{\main/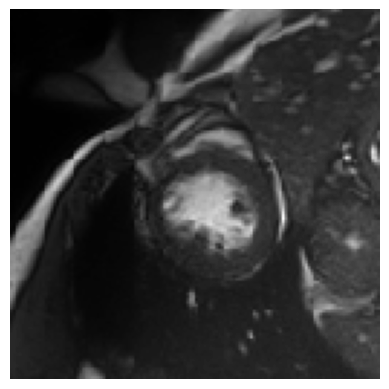}};
    \node[draw=white, inner sep=0] (img1) at (-2.3,0.2) {\includegraphics[height=1.0cm]{\main/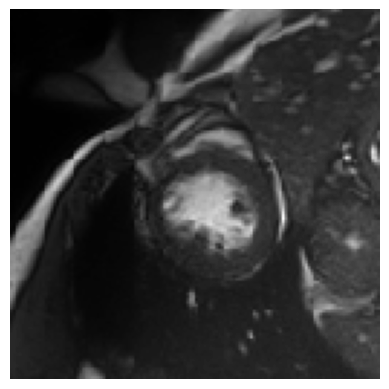}};
    \node[draw=white, inner sep=0] (img2) at ([xshift=5pt,yshift=-5pt]img1) {\includegraphics[height=1.0cm]{\main/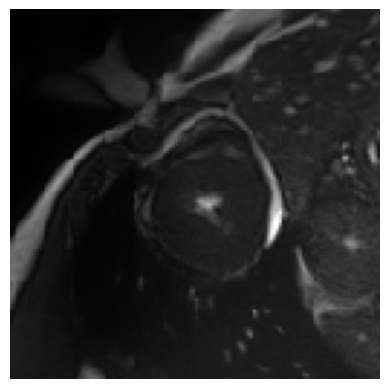}};
    \node[draw=white, inner sep=0] (img3) at ([xshift=5pt,yshift=-5pt]img2) {\includegraphics[height=1.0cm]{\main/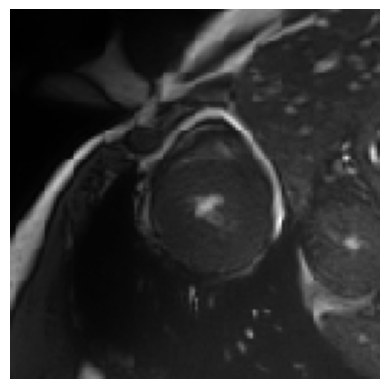}};
    \node[draw=white, inner sep=0] (img4) at ([xshift=5pt,yshift=-5pt]img3) {\includegraphics[height=1.0cm]{\main/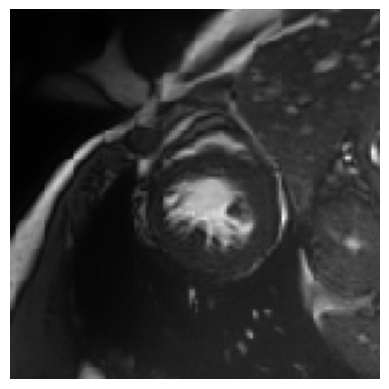}};
    \node[label=below:{\Large $\seq{y}$}] (img1T) at (-2,-0.7) {};

    \node[const] at (0,0) (enc_lb) {}; 
    \node[const] at (0,3) (enc_lt) {}; 

    \node[const] at (4,2) (enc_rt) {}; 
    \node[const] at (4,1) (enc_rb) {}; 

    \node[const] at ($(enc_rt)!0.5!(enc_rb)$)  (enc_r) {};
    \node[const] at ($(enc_lt)!0.5!(enc_lb)$)  (enc_l) {};
    \node[const] at ($(enc_l)!0.3!(enc_r)$)  (enc_m) {};

    \node[const] at ($(enc_lt)!0.4!(enc_rt)$)  (x1) {};
    \node[const] at ($(enc_lt)!0.6!(enc_rt)$)  (x2) {};
    \node[const] at ($(enc_lt)!0.8!(enc_rt)$)  (x3) {};

    \draw[semithick] (enc_rb) -- (enc_lb);
    \draw[semithick] (enc_rt) -- (enc_lt);
    \draw[semithick] (enc_lb) arc (270:90:0.5 and 1.5);
    \draw[semithick] (enc_r) ellipse (0.166 and 0.5);

    \draw (enc_m) node {\makecell[c]{
        \Large $q_\phi(x_t \mid y_0, y_t)$ 
    }};


    \node[const] at (7,1) (dec_lb) {}; 
    \node[const] at (7,2) (dec_lt) {}; 

    \node[const] at (11,0) (dec_rb) {}; 
    \node[const] at (11,3) (dec_rt) {}; 

    \node[const] at ($(dec_rt)!0.5!(dec_rb)$)  (dec_r) {};
    \node[const] at ($(dec_lt)!0.5!(dec_lb)$)  (dec_l) {};
    \node[const] at ($(dec_l)!0.55!(dec_r)$)  (dec_m) {};

    \draw[semithick] (dec_lb) arc (270:90:0.166 and 0.5);
    \draw[semithick] (dec_rb) -- (dec_lb);
    \draw[semithick] (dec_rt) -- (dec_lt);
    \draw[semithick] (dec_r) [partial ellipse=300:60:0.5 and 1.5];

    \node[const] at ($(dec_lt)!0.2!(dec_rt)$)  (x13) {};
    \node[const] at ($(dec_lt)!0.4!(dec_rt)$)  (x12) {};
    \node[const] at ($(dec_lt)!0.6!(dec_rt)$)  (x11) {};

    \node[const] at ($(enc_r)!0.5!(dec_l)$) (encdec_m) {};

    \node [draw, semithick, circle] at (encdec_m) (latent) {\makecell[c]{
        \Large $\mu_t^{\text{enc}}$\\ \\
        \Large $\Sigma_t^{\text{enc}}$}};

    \node[const, right= 0.5cm of dec_rt] (w_lt) {};
    \node[const, right= 1cm of w_lt] (w_rt) {};

    \node[const, right= 0.5cm of dec_rb] (w_lb) {};
    \node[const, right= 1cm of w_lb] (w_rb) {};

    \node[const, right=-1.7cm of dec_m] (phi) {\makecell[c]{        
        \Large $\varphi_t = g_\theta(x_t, s)$ 
    }};

    \node[const] at ($(w_lt)!0.5!(w_rt)$)  (w_mt) {};
    \node[const] at ($(w_lb)!0.5!(w_rb)$)  (w_mb) {};
    \node[const] at ($(w_lt)!0.5!(w_lb)$)  (w_l) {};
    \node[const] at ($(w_rt)!0.5!(w_rb)$)  (w_r) {};
    \node[const] at ($(w_l)!0.1!(w_r)$)  (w_m) {};

    \draw[semithick] (w_lb) arc (270:90:0.5 and 1.5);
    \draw[semithick] (w_lb) -- (w_rb);
    \draw[semithick] (w_lt) -- (w_rt);
    \draw[semithick] (w_r) ellipse (0.5 and 1.5);
    \draw (w_m) node[rotate=90] {\makecell[c]{\Large $p_\theta(y_t \mid y_0, \varphi_t)$}};    

    \node[const, below = 1.0cm of latent] (eq) {%
    \makecell[c]{        
        \Large LG-SSM
    }};

    \node[const] at ($(dec_l)!0.50!(w_r)$)  (dec_c) {};
    \node[const] at ($(enc_l)!0.5!(enc_r)$)  (enc_c) {};

    \plate[black, fill=red, fill opacity=0.2, rounded corners,fit margins={left=10pt,right=5pt,bottom=2pt,top=2pt}] {encoder} {(enc_rt)(enc_rb)(enc_lt)(enc_lb)} {\Large $\phi$} ; %
    \node[const, below=0.4cm of enc_c] {\includegraphics[width=0.10\textwidth]{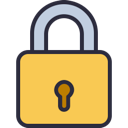}};

    \node[const, above=0.2cm of dec_m, yshift=1.7cm]  (fs) {\makecell[c]{\Large $f^s_\theta(y_0)$}};    
    \plate[rounded corners, fit margins={left=15pt,right=15pt,bottom=0pt,top=2pt}] {fsbox} {(fs.north west)(fs.south west)(fs.north east)(fs.south west)} {} ; %
    \coordinate (C13) at ($(fsbox.north)!1!(x13)$);
    \node[const] at (C13 |- 2,3.2) (s13) {};    
    \path (s13) edge [-triangle 45] node {} (x13);

    \coordinate (C11) at ($(fsbox.north)!1!(x11)$);
    \node[const] at (C11 |- 2,3.2) (s11) {};    
    \path (s11) edge [-triangle 45] node {} (x11);

    \plate[black, fill=green, fill opacity=0.2, rounded corners, fit margins={left=5pt,right=10pt,bottom=3pt,top=4pt}] {generative} {(fs.north west)(w_rb)(fs.north east)(dec_lb)} {\Large $\theta$} ; %
    \draw[->] (imgM.north) -- (imgM.north |- fsbox.west) -- (fsbox.west);

    \node[const, below=0.4cm of dec_c] {\includegraphics[width=0.10\textwidth]{128x128.png}};

    \plate[black, fill=LightBlue, fill opacity=1.0, rounded corners,fit margins={left=2pt,right=2pt,bottom=0pt,top=4pt}] {lgssm} {(eq.north west)(eq.south west)(eq.north east)(eq.south east)} {\Large $\gamma$};

    \node[const, below = 1.0cm of latent] (eq1) {%
    \makecell[c]{        
        LG-SSM \eqref{eq:lgssm} \\
        \large $\gamma$
    }};
    \path (latent.south) edge [-triangle 45] node {} (lgssm.north);
\end{tikzpicture}

%% file: 04_Experiment.tex
\section{Experiments}
For experiments, we evaluate our model on two publicly available datasets: i) single-cycle cine-MRI sequences from the Automatic Cardiac Diagnosis Challenge (ACDC)~\cite{bernard2018deep}, and ii) longer sequences of cardiac ultrasound images from the EchoNet-Dynamic database~\cite{ouyang2020video}. On the EchoNet-Dynamic dataset, we perform online learning with a moving horizon of $N=75$. This is not suitable for the ACDC dataset as the sequences are too short. Instead, on the ACDC dataset we evaluate the capability to reconstruct the sequence from sparsely sampled sequences where we only observe a subset of the images in the sequence.

\begin{figure}
    \centering
    \begin{tabular}{ccccc|c}
        \footnotesize $t_{T+1}=176$ & \footnotesize $t=188$ & \footnotesize $t=201$ & \footnotesize $t=213$ 
        & \footnotesize $t=225$ & \footnotesize Dice  \\        
        \includegraphics[scale=0.25]{\main/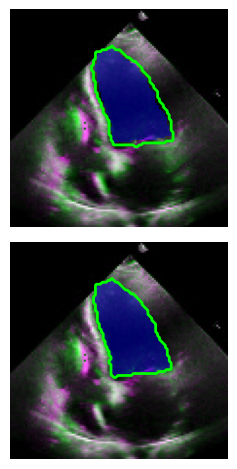} & 
        \includegraphics[scale=0.25]{\main/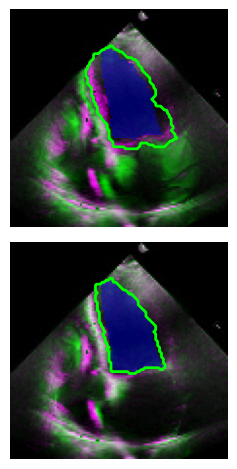} & 
        \includegraphics[scale=0.25]{\main/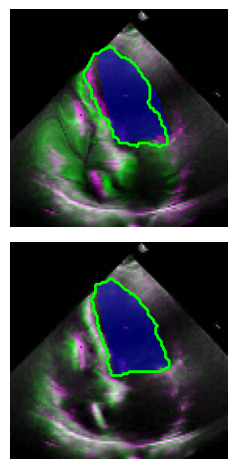} & 
        \includegraphics[scale=0.25]{\main/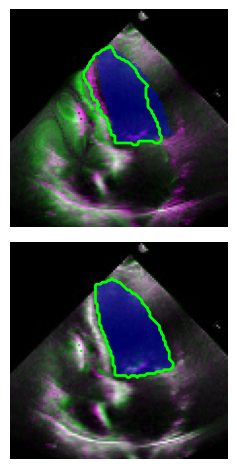} &         
        \includegraphics[scale=0.25]{\main/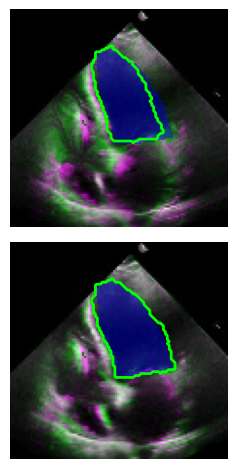} &
        \includegraphics[scale=0.25]{\main/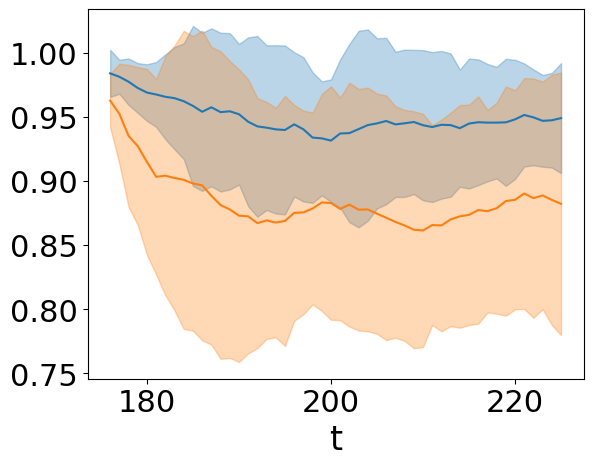}\\ 
        \multicolumn{5}{c|}{
        \tikz\draw[magenta,fill=magenta] (0,0) circle (.3ex); \footnotesize $y_t$  \quad \tikz\draw[green,fill=green] (0,0) circle (.3ex); \footnotesize $y_0 \circ \varphi_t$  \quad  \tikz \node[rectangle, draw, blue,fill=blue] (0,0) (.3ex) {};  \footnotesize $t\mid t$  \quad \tikz \node[rectangle, draw, green] (0,0) (.3ex) {}; \footnotesize $t\mid t_T$ (forecast)} & \footnotesize \myorange{Pre-trained} vs. \myblue{Online}\\        
    \end{tabular}%
    \caption{Overlay between true sequence (magenta) and forecasted sequence (green) using pre-trained model, on top, and online learning, on bottom. To the right, Dice score distribution of the left ventricle from 20 forecasted samples and the estimated region given the entire sequence. The figure is colored in the online version. }
    \label{fig:echoresult}%
\end{figure}

Both datasets are segmented manually at the end-systole and the end-diastole time points. We use the first time point as our static reference image and calculate the Dice score coefficients~(DSC) and the $95\%$-th Hausdorff distance (HD95) between the other manually segmented frame and our estimation at the given point for evaluation. We compare our registration accuracy against no estimated motion and two well-established image registration methods: symmetric normalization (SyN)~\cite{avants2008symmetric} and elastic registration~\cite{modersitzki2003numerical}, both using the ANTs software~\cite{avants2009advanced}. Moreover, for online learning, we leave a horizon of $H=50$ samples for each sequence and calculate both the log-likelihood of the unseen sequence $x_{T:T+H}$ and the RMSE between $50$ samples from the forecasting distribution and the true latent values. Furthermore, we also calculated the Dice score between the samples $25$ steps ahead and the estimated segmentation given the entire sequence. In Table~\ref{tab:result}, we present the overall result from both datasets. All models produce diffeomorphic deformations (positive Jacobian determinants) and this metric is omitted from the table. The execution times for motion estimation with and without online learning are approximately $15$~ms and $75$~ms on a single CPU, respectively.
\begin{figure}
    \begin{tabular}{cacac|c}        
        \footnotesize $t=5$ & \footnotesize $t=10$ & \footnotesize $t=15$ & \footnotesize $t=20$ & \footnotesize $t=25$
        &  \\        
        & \footnotesize obs 
        &
        &\footnotesize obs   &
        & \footnotesize LV cm$^2$ \\        
        \includegraphics[scale=0.25]{\main/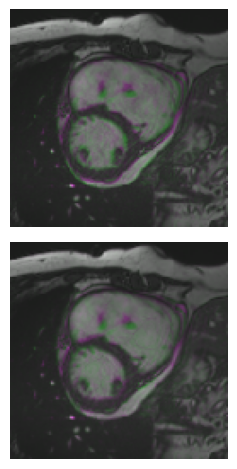} &         
        \includegraphics[scale=0.25]{\main/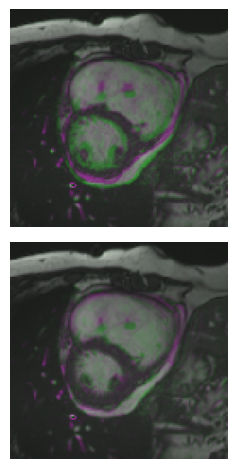} &         
        \includegraphics[scale=0.25]{\main/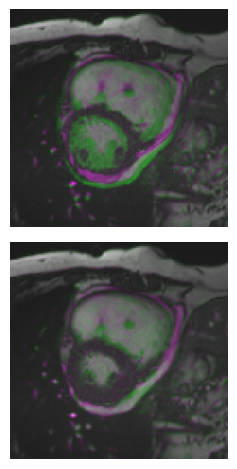} & 
        \includegraphics[scale=0.25]{\main/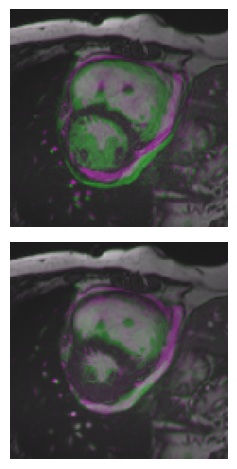} & 
        \includegraphics[scale=0.25]{\main/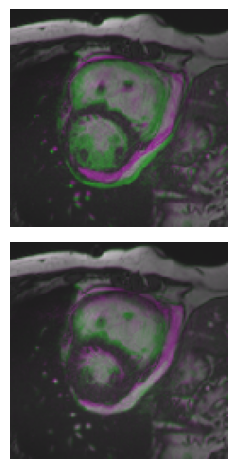} &
        \includegraphics[scale=0.25]{\main/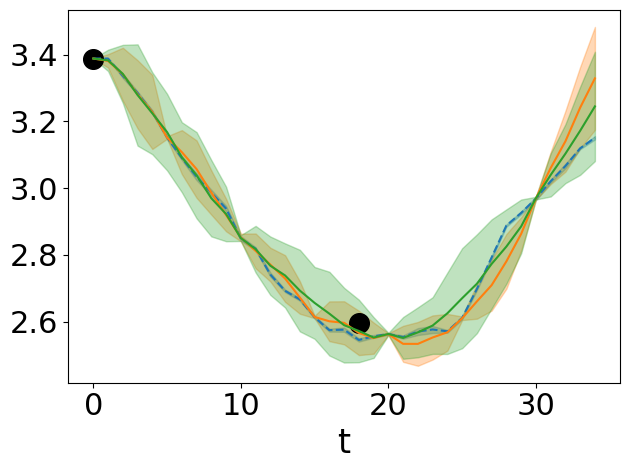}             
         \\
        \multicolumn{5}{c|}{
        \tikz\draw[magenta,fill=magenta] (0,0) circle (.3ex); \footnotesize $y_t$ 
        \tikz\draw[green,fill=green] (0,0) circle (.3ex); \footnotesize $y_0 \circ \varphi_t$} 
        & \tikz\draw[black,fill=black] (0,0) circle (.3ex); \footnotesize GT
          \tikz\draw[LightBlue,fill=LightBlue] (0,0) circle (.3ex); \footnotesize All 
          \tikz\draw[LightOrange,fill=LightOrange] (0,0) circle (.3ex); \footnotesize 5th 
          \tikz\draw[LightGreen,fill=LightGreen] (0,0) circle (.3ex); \footnotesize 10th
    \end{tabular}
    \caption{Overlay of true sequence (magenta), and $\varphi_t = 0$, on top, and our estimation given every 10th sample, on bottom (green). On the right, the distribution of the left ventricle area for $20$ latent samples under three scenarios: all time points observed, every 5th, and every 10th. The figure is colored in the online version.}
    \label{fig:acdcresult}%
\end{figure}
\\ \\
\textbf{ACDC:} In the ACDC experiment, which consists of 100 patients for training and 50 for testing, we resample the images with spacing $1.5 \times 1.5$ mm and crop it to $128 \times 128$ pixels with the ventricles in the center. The original sequences are in 3D with limited resolution in one orientation. Therefore we only consider the 2D motion in the other two orientations. For training, we split the volume into slices and removed slices with no annotations, resulting in a training set of 840 sequences. In the evaluation part, we use the middle slice of the volume in the test dataset to ensure connected segmented regions with no mismatch due to out-of-plane motions. For consistency regarding the sequence length, we resample each sequence to $35$ samples using bilinear interpolation. During the training phase, we augment the data using random rotation, flip, and translation of the whole sequences. For evaluation, we use the segmented regions of the right ventricle (RV), the left ventricle myocardium (LV-Myo), and the left ventricle blood pool (LV-BP). Fig.~\ref{fig:acdcresult} shows the result from where we reconstruct the entire sequence using only every $10$th sample as input to our model.
\\ \\
\textbf{EchoNet-Dynamic:} The EchoNet-Dynamic dataset includes $10\,023$ unique cardiac ultrasound videos of various lengths with left ventricle segmentations~(LV). We split this data into a training set of $9\,540$ videos and $483$ videos for testing. Furthermore, during training, for each epoch, we randomly selected a sequence of $50$ frames from each video. Fig.~\ref{fig:echoresult} shows the result of online training when we forecast the motion $50$ time-steps ahead.
\begin{table}
    \caption{Overall results from the two datasets.}
    \label{tab:result}%
    \centering
        \begin{tabular}{l|cccccccc}
        \toprule
        & \multicolumn{6}{c}{\textbf{ACDC}}  & \multicolumn{2}{|c}{\textbf{EchoNet}} \\ 
        & \multicolumn{3}{c}{DSC} &  \multicolumn{3}{c}{HD95[mm]} & \multicolumn{1}{|c}{DSC} &  HD95$^1$\\  
        & RV & LV-Myo & LV-BP & RV & LV-Myo & LV-BP &\multicolumn{1}{|c}{LV} & LV \\
        \midrule
        None & $0.70$ & $0.52$ & $0.66$ & $9.11$ & $8.46$ & $11.06$ & \multicolumn{1}{|c}{$0.74$} & $8.37$\\ 
        Elastic & $0.77$ & $0.72$ & $0.81$ & $6.81$ & $5.88$ & $6.33$ & \multicolumn{1}{|c}{$0.87$} & $5.94$\\
        SyN & $0.79$ & $0.72$ & $\mathbf{0.86}$ & $5.88$ & $5.22$ & $\mathbf{4.69}$  & \multicolumn{1}{|c}{$\mathbf{0.88}$} & $\mathbf{4.68}$\\
        Our & $\mathbf{0.80}$ & $\mathbf{0.82}$ & $0.84$ & $5.35$ & $\mathbf{4.34}$ & $5.18$ & \multicolumn{1}{|c}{$0.86$} & $4.92$\\
        Our $5$th & $0.80$ & $0.81$ & $0.84$ & $\mathbf{5.28}$ & $4.44$ & $5.27$ & \multicolumn{1}{|c}{$-$}& $-$\\ 
        Our $10$th & $0.79$ & $0.80$ & $0.83$ & $5.51$ & $4.63$ & $5.57$& \multicolumn{1}{|c}{$-$}& $-$\\ 
        \midrule
        & \multicolumn{8}{c}{\textbf{EchoNet Forecasting}} \\
        & \multicolumn{2}{c}{$\log p_\gamma(x_{T:T+H})$} & 
        & \multicolumn{2}{c}{RMSE ($x_{T:T+H}$)}  &
        & \multicolumn{2}{c}{DSC ($\varphi_{T+25 \mid T}$)} \\                
        \midrule
        Pre-trained & \multicolumn{2}{c}{$-10.5$} & &\multicolumn{2}{c}{$7.04$} & & \multicolumn{2}{c}{$0.81$} \\
        Online & \multicolumn{2}{c}{$\mathbf{-6.3}$} &  & \multicolumn{2}{c}{$\mathbf{5.54}$} & & \multicolumn{2}{c}{$\mathbf{0.85}$}\\
        \bottomrule
    \end{tabular}
    \\
    \footnotesize{$^1$ Spacing is not specified in the dataset. The metric is given in pixels.}
\end{table}

%% file: 05_Discussion.tex
\section{Discussion and conclusion}
In this work, we have presented a motion model for intra-interventional medical images. We define the motion model in a low-dimensional space as a probabilistic LG-SSM with analytical solutions to the inference problem, like imputation for undersampled data (smoothing) and forecasting into the future (prediction). In the first experiment, on the ACDC dataset, we show a marginally improved accuracy compared to well-established diffeomorphic image registration methods, even in cases where we subsample the data and retain only $10\%$ of the original sequence. Our model, operating in a lower and more manageable latent space, shows similar accuracy to recent work~\cite{krebs2021learning}. However, a direct comparison is not feasible since both the code and some of the data are not publicly available. In the second experiment, on the EchoNet-dynamics dataset, we show the capacity of the model to adapt to new, patient-specific data by using online learning and updating the weights in the low-dimensional LG-SSM. The online learned model shows forecasting improvements in both similarities of the latent process with higher likelihood given the true process and lower distance between the samples compared to the pre-trained model as well as the calculated Dice score for predicted samples. The registration accuracy in this experiment is slightly worse than the conventional image registration methods, and can hopefully be improved by refining the hyperparameter settings or data preprocessing. Finally, we believe patient-specific adaptation and reliable forecasting predictions are necessary for longer sequences to support advanced procedures, like real-time adaptation in MR-guided radiotherapy. Other topics for further investigation include relating the uncertainty in the latent temporal process to the uncertainty in the estimated displacement field and observing how each component of the latent space contributes to the actual motion.

%% file: SupplementaryFile.tex
\institute{}
\begin{center}
\textbf{\Large Supplemental Materials:\\
\normalsize Online learning in motion modeling for intra-interventional image sequences}
\end{center}

\setcounter{equation}{0}
\setcounter{figure}{0}
\setcounter{table}{0}
\setcounter{page}{1}
\subsection*{Derivation of the ELBO}
The conditional probability density function
\begin{align}
    p(\seq{y} \mid y_0) = \dfrac{p(\seq{x,y,z}\mid y_0)}{p(\seq{x,z}\mid y_0, \seq{y})},
\end{align}
is infeasible due to the intractable posterior distribution $p(\seq{x,z}\mid y_0, \seq{y})$. Instead, we can approximate the posterior distribution, and identify a lower bound of $p(\seq{y} \mid y_0)$. In KVAE the posterior distribution is approximated as
\begin{align}
    q(\seq{x,z} \mid y_0, \seq{y}) = q_\phi(\seq{x} \mid y_0, \seq{y}) p_\gamma(\seq{z} \mid \seq{x}),
\end{align}
where $q_\phi(\seq{x} \mid y_0, \seq{y}) = \prod_{t=1}^T q_\phi(x_t \mid y_0, y_t)$ is parameterized using the inference network, i.e.
\begin{align}
    q_\phi(x_t \mid y_0, y_t) = \mathcal{N}(x_t \mid \mu^{\text{enc}}_t, \Sigma^{\text{enc}}_t).
\end{align}
If we rewrite the true posterior distribution 
\begin{align}
    p(\seq{x,z}\mid y_0, \seq{y}) = \dfrac{p(y_0, \seq{y,x,z})}{p(y_0, \seq{y})},
\end{align}
and derive the full distribution model
\begin{align}
    p(y_0, \seq{y, x, z}) = p(y_0) p_\theta(\seq{y} \mid y_0, x) p_\gamma(\seq{x,z}),
\end{align}
the true posterior distribution is equivalent to
\begin{align}
    p(\seq{x,z}\mid y_0, \seq{y}) 
    & = \dfrac{p(y_0)p_\theta(\seq{y} \mid y_0, \seq{x})p_\gamma(\seq{x}, \seq{z})}{p(y_0, \seq{y})} \\
    &  = \dfrac{p_\theta(\seq{y} \mid y_0, \seq{x})p_\gamma(\seq{x}, \seq{z})}{p(\seq{y} \mid y_0)}.
\end{align}
Next, from the KL divergence between the true posterior distribution and our approximate posterior distribution
\begin{align}
    \text{D}_{\text{KL}}(q(\seq{x,z} \mid y_0, \seq{y}) || p(\seq{x,z} \mid y_0, \seq{y})) \geq 0,
\end{align}
we have that 
\begin{align}
    \text{D}_{\text{KL}}(q || p) = 
    & \mathbb{E}_{q(\seq{x,z} \mid y_0, \seq{y})} \left[ \log 
    \dfrac{
        q(\seq{x,z} \mid y_0, \seq{y})
    }{
        p(\seq{x,z} \mid y_0, \seq{y})
    }\right] \\
    = & \mathbb{E}_{q(\seq{x,z} \mid y_0, \seq{y})} \left[ \log 
    \dfrac{
        q_\phi(\seq{x} \mid y_0, \seq{y})
        p_\gamma(\seq{z} \mid \seq{x})
        p(\seq{y} \mid y_0)
    }{
        p_\theta(\seq{y} \mid y_0, \seq{x})
        p_\gamma(\seq{x}, \seq{z})
    }\right] \\
    = & \log p(\seq{y} \mid y_0) - \mathbb{E}_{q(\seq{x,z} \mid y_0, \seq{y})} \left[
        \log \dfrac{
            p_\theta(\seq{y} \mid y_0, \seq{x})
            p_\gamma(\seq{x}, \seq{z})
        }{
            q_\phi(\seq{x} \mid y_0, \seq{y})
            p_\gamma(\seq{z} \mid \seq{x})
        }
    \right]
    \geq 0
\end{align}
Finally, by moving the expectation to the right-hand side of the inequality, a tractable lower bound of the likelihood is identified
\begin{align}
    \log p(\seq{y} \mid y_0) \geq \mathbb{E}_{q(\seq{x,z} \mid y_0, \seq{y})} \left[
        \log \dfrac{
            p_\theta(\seq{y} \mid y_0, \seq{x})
            p_\gamma(\seq{x}, \seq{z})
        }{
            q_\phi(\seq{x} \mid y_0, \seq{y})
            p_\gamma(\seq{z} \mid \seq{x})
        }
        \right].
\end{align}
\subsection*{Model architecture}
\textbf{Encoder} ($129\text{k}$ \& $84\text{k}$ parameters): 
The inference network and the spatial feature extraction share a similar network architecture. We downsample the data using a stack of convolutional layers, where we extract spatial features at each resolution. The network downsamples the data four times using CNNs with filters $[32, 32, 32, 16]$ and then flattens and feeds the features into a dense network. We approximate the posterior distribution by estimating the mean and covariance of $x_t$. \\ \\
\textbf{Decoder} ($129\text{k}$ parameters):
For the generative network, we use attention gates to focus the temporal changes on the spatial features of the reference image at each resolution, followed by an upsampling CNN. The upsampling uses the same number of resolution layers and filters per level as the downsampling. At the output level, we apply a Gaussian filter (with $\sigma_G = 2$ in the ACDC model and $\sigma_G = 4$ in the EchoNet-Dynamic model) after the last convolutional layer. To enforce diffeomorphic estimates of $\varphi_t$, we consider the output as the stationary velocity field $v_t$ and compute the transformation numerically using four scaling and squaring layers. \\ \\
\textbf{LG-SSM} ($976$ parameters):
We design the LG-SSM using eight dimensions for $x_t$ ($p=8$) and $16$ for the state-variable $z_t$ ($q=16$). We estimate the full matrices $A, C$, the initial mean $\mu_0$, and the lower triangular matrices of the covariances $R, Q, \Sigma_0$. \\ \\
\textbf{Training procedure:}
For training purposes, we transform the reference image $y_0$ using the estimated spatial transformation to compute the likelihood $p_\theta(y_t \mid y_0, \varphi_t)$. For the ACDC experiment, we use a local cross-correlation distribution as likelihood and a Gaussian distribution in the EchoNet-dynamic experiment. We optimize the network using Adam optimizer with a learning rate $5\times 10^{-4}$ in both the offline and online scenarios. During offline training, we used a batch size of $4$ and trained the ACDC model for $500$ epochs and the EchoNet-Dynamic model for $50$ epochs.
\newpage